\documentclass[sigconf]{acmart}
\renewcommand\footnotetextcopyrightpermission[1]{} % removes footnote with conference information in first column
\usepackage{booktabs} % For formal tables

\usepackage{algorithm} % algorithm
\usepackage{algorithmic} % algorithm
\usepackage{mdwlist} % for itemize
\usepackage[mathscr]{euscript} % for euler script.
\usepackage{amsbsy} % for \boldsymbol
\usepackage{amsfonts} % for mathbb
\usepackage{subfig} % for \subfloat
\usepackage{array} % for \newcolumntype
\usepackage{ragged2e} % for right align in table

\usepackage{graphicx}
\usepackage{amsmath} % for pmatrix
\usepackage{hyperref} % for \url
\usepackage{multirow} % for table
\usepackage{footnotebackref}[2012/07/01]% v1.0
\usepackage{tablefootnote}[2014/01/26] % for table footnote
\usepackage{color}
\usepackage{xspace}

\usepackage{url}

\setcopyright{none}
\newcommand{\T}[1]{\boldsymbol{\mathscr{#1}}}

\newcommand{\bit}{\begin{itemize*}}
	\newcommand{\eit}{\end{itemize*}}

\newcommand{\hide}[1]{}

\newcolumntype{C}[1]{>{\centering\arraybackslash}p{#1}}
\newcolumntype{R}[1]{>{\RaggedLeft\arraybackslash}p{#1}}
\newcolumntype{L}[1]{>{\RaggedRight\arraybackslash}p{#1}}

\newcommand{\algorithmicdoinparallel}{\textbf{do in parallel}}
\makeatletter
\AtBeginEnvironment{algorithmic}{%
	\newcommand{\FORP}[2][default]{\ALC@it\algorithmicfor\ #2\ %
		\algorithmicdoinparallel\ALC@com{#1}\begin{ALC@for}}%
	}
	\makeatother

\begin{document}
\title{SNeCT: Scalable network constrained Tucker decomposition for integrative multi-platform data analysis}

\author{Dongjin Choi}
\affiliation{%
	\institution{Seoul National University}
}
\email{skywalker5@snu.ac.kr}

\author{Lee Sael}
\affiliation{%
	\institution{The State University of New York (SUNY) Korea}
}
\email{sael@sunykorea.ac.kr}
%\author{Anonymous}
%\affiliation{\institution{Anonymous}}
%% The default list of authors is too long for headers}
%\renewcommand{\shortauthors}{Anonymous}

\begin{abstract}
    \textbf{Motivation:}
How do we integratively analyze large-scale multi-platform genomic data that are high dimensional and sparse?
Furthermore, how can we incorporate prior knowledge, such as the association between genes, in the analysis systematically?\\
\textbf{Method:} To solve this problem, we propose a \textbf{S}calable \textbf{Ne}twork \textbf{C}onstrained \textbf{T}ucker decomposition method we call SNeCT.
SNeCT adopts parallel stochastic gradient descent approach on the proposed parallelizable network constrained optimization function.
SNeCT decomposition is applied to tensor constructed from large scale multi-platform multi-cohort cancer data, PanCan12, constrained on a network built from PathwayCommons database.\\
\textbf{Results:} The decomposed factor matrices are applied to stratify cancers, to search for top-$k$ similar patients, and to illustrate how the matrices can be used for personalized interpretation.
In the stratification test, combined twelve-cohort data is clustered to form thirteen subclasses.
The thirteen subclasses have a high correlation to tissue of origin in addition to other interesting observations, such as clear separation of OV cancers to two groups, and high clinical correlation within subclusters formed in cohorts BRCA and UCEC.
In the top-$k$ search, a new patient's genomic profile is generated and searched against existing patients based on the factor matrices.
The similarity of the top-$k$ patient to the query is high for 23 clinical features, including estrogen/progesterone receptor statuses of BRCA patients with average precision value ranges from 0.72 to 0.86 and from 0.68 to 0.86, respectively.
We also provide an illustration of how the factor matrices can be used for interpretable personalized analysis of each patient. \\
\textbf{Availability:} The code and data available at our repository\footnote{https://github.com/skywalker5/SNeCT}.\\
\textbf{Supplementary information:} Supplementary data are available at \textit{Bioinformatics}
online.
\end{abstract}

%
% The code below should be generated by the tool at
% http://dl.acm.org/ccs.cfm
% Please copy and paste the code instead of the example below.
%
%\begin{CCSXML}
%<ccs2012>
%<concept>
%<concept_id>10002951.10003227.10003351</concept_id>
%<concept_desc>Information systems~Data mining</concept_desc>
%<concept_significance>300</concept_significance>
%</concept>
%%	<concept>
%%	<concept_id>10003752.10003809</concept_id>
%%	<concept_desc>Theory of computation~Design and analysis of algorithms</concept_desc>
%%	<concept_significance>500</concept_significance>
%%	</concept>
%</ccs2012>
%\end{CCSXML}
\maketitle
%\vspace*{-0.3cm}
\section{Introduction}
    \label{sec:intro}
    \begin{figure*} [!t]
	\begin{center}
		\includegraphics[width=1 \textwidth]{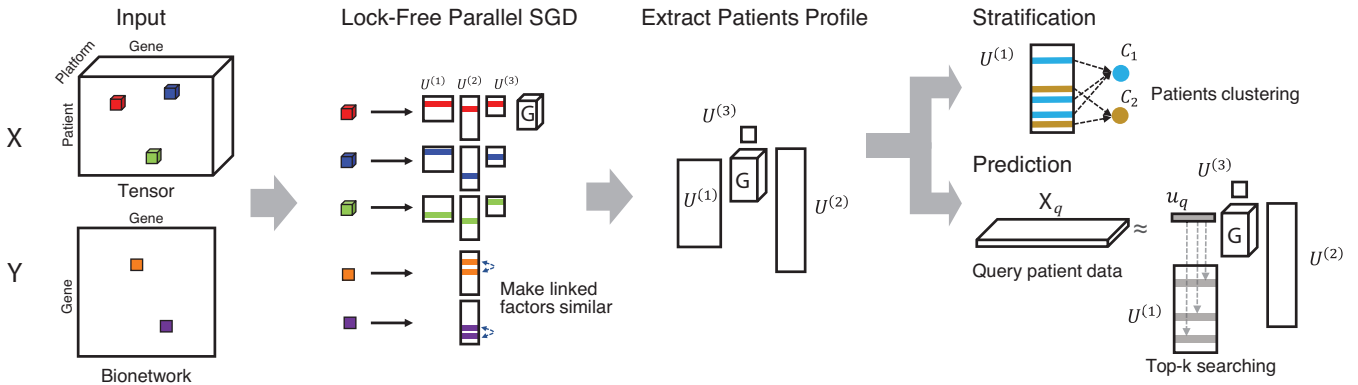}
		%		\vspace*{-0.5cm}
		\caption{Overview of the tensor decomposition of SNeCT and validation processes.}
		\label{fig:overview}
	\end{center}
	%	\vspace*{-0.5cm}
\end{figure*}

% Importance of integrative analysis of bio-data
Integrative analysis of multiple perspectives of a patient helps in both stratification and clinical predictions.
Stratification helps the researchers in understanding and exploring the genomic characteristics in relation to their current phenotypes and thus to recognize opportunities for clinical improvement on stratified groups of patients.
In the perspective of personalized medicine, clinical diagnostics and predictions of individual patient is needed and can be done by searching the integrated profile of a patient to existing records.
Analysis of one or few data types may not be sufficient for stratifications or accurate predictions of disease as they only provide partial information about the patient's biological status.
%In cancer data analysis, an improved stratification, and clinical prediction can be achieved by integrative analysis of the multi-platform data such as copy number variation (CNV), somatic mutation, gene expression, DNA methylation, and microRNA (miRNA) data.
%The Genomic Data Commons (GDC) Data Portal \href{https://gdc.cancer.gov/}{https://gdc.cancer.gov/} reports diverse genomic information with paired clinical information.

% Survey of integrative analysis methods:
% copied from previous paper; need to update the short review.
\textbf{Related works in integrative genomic analysis: }
Need for integrative data analysis methods is being recognized, however, due to increased sized of data and a limited number of uniform data analysis framework, integrative analysis of multiple data types is still a challenging task.
Existing methods are often limited in interpretability and scalability and often runs in selected subset of data and features.
Other methods run only on small number of genes such as work by \cite{Louhimo11} that shows the effect of DNA methylation and CNV in gene expression of several known oncogenes for glioblastoma and ovarian cancer;
PARADIGM method by \cite{Vaske2010} that adopts graph inference approach on augmented pathway structure containing nodes for CNV, gene expression, protein expression and active protein information; and a work by \cite{Sohn13} that proposes an integrative statistical framework based on a sparse regression of gene expression values based on CNV, miRNA, and methylation.
Many methods run only on small data sets, such as multiple-kernel based method by \cite{Thomas2015, Thomas2016b} that combines kernels generated from individual platform data in a weighted linear fashion for stratification and predictions of ovarian cancer; a method by \cite{Mankoo11} that applies multivariate Cox Lasso model and median time-to-event prediction algorithm on dataset integrated from the CNV, methylation, miRNA, and gene expression data; and iCluster method by \cite{Mo2013} that transforms multi-platform data to latent space and clusters using latent variable.
Also, many methods do not truly integrate the data in the analysis such as work by \cite{Yuan14} that evaluated the predictive power of patient survival and clinical outcome using clinical data in combination with one of CNV, methylation, mRNA, miRNA or protein expression data, and work by \cite{Hoadley2014} that stratifies the PANCAN12 dataset using multi-platform data where clustering results of each data type are used to construct the final cluster.

\textbf{Multi-platform multi-cancer data analysis: }
% scalable, interpretable, integrative method needed.
The data scalability challenge in the integrative analysis is even more evident in multi-platform data analysis across multiple cancer types.
Analysis across multiple cancer types enables us to get a glimpse of the extent to which genomic signatures are shared across the different cancers.
The biological understanding of similarity and dissimilarity among the different cancer types can enable efficient management of diseases as well as treatment transfers between different cancer types of similar genomic signatures.
The work of \cite{Hoadley2014} is one of the first attempts to utilize multi-platform data of multiple cancers, i.e. the PanCan12 dataset.
The PanCan12 dataset is created by the Pan-Cancer initiative that compares 12 tumor types profiled by The Cancer Genome Atlas (TCGA) Research Network and includes data from six different platforms \cite{Chang2013}.
In their work, integrated subtype classification for all of the tumor samples was performed by first clustering on individual data platforms, and then using the results of single-platform clusters as input to a second-level cluster analysis to form a cluster-of-cluster assignment (COCA).
Fully integrative data analysis method would consider and optimize the multi-platform data altogether. In this perspective, the COCA method can be considered as an ensemble method where the input data are varied, rather than an integrative method.
Also, it is difficult to utilize the COCA approach for finding similar patients as needed in the clinical predictions given a new patient's data without redoing the analysis over again. Thus, there is a need for multi-platform data analysis method that can scalably stratify multiple cancer types for knowledge discovery and predict clinical outcomes for enabling personalized medicine.

\textbf{Related works in tensor analysis: }
% review on tensor; can add more reference if needed.
Tensors, i.e., multi-dimensional arrays, are a natural representation of multi-platform genomic data \cite{Sael2015}.
The core of tensor analysis is tensor decomposition, which can be considered as higher-order singular value decomposition (HOSVD).
Tensor analysis has been widely applied with success on network traffic \cite{Maruhashi2011},
knowledge bases \cite{Carlson2010, Nickel2012},
hyperlinks and anchor texts in the Web graphs \cite{Kolda2006},
sensor streams \cite{Sun2006},
and DBLP conference-author-keyword relations \cite{Kolda2008}, to name a few.
The major challenge of tensor analysis is data scalability as there is an intermediate data explosion problem in the decomposition process even when the input tensor fits into the memory.
To address this problem we have previously proposed a Hadoop based parallel tensor decomposition method \cite{Jeon2016, Jeon2016a} and
a multi-core based method \cite{Shin2017}.

\textbf{Contributions: }
In this paper, we propose a tensor-based method that enables stratification and clinical prediction of patients utilizing multi-platform data analysis across multiple cancer types.
We have shown in our previous works \cite{Kim2015c, Kim2014} that somatic mutation profiles generated from orthogonal matrix decomposition enable accurate stratification and clinical predictions of each cancer type.
We extend this approach to multi-platform multi-cancer data analysis by proposing a scalable network constrained Tucker decomposition (SNeCT) method (Figure~\ref{fig:overview}), and show that SNeCT can efficiently stratify cancer subtypes and predict clinical outcomes. The contributions of this paper are listed in the following.
\begin{itemize}
	\item A novel scalable network constrained Tucker decomposition (SNeCT) algorithm.
	\item Stratification on multi-platform multi-cohort data showing similarity and difference between cancer cohorts.
	\item Individualized clinical prediction utilizing multi-platform genomic profiles.
	\item A demonstration of personalized interpretation utilizing factor matrices.
\end{itemize}
%\vspace{-0.2cm}
%\vspace*{-0.3cm}

\section{Materials and Methods}
\label{sec:method}
    \begin{table}[bp]
	\small
	\caption{TCGA PAN Cancer (PanCan12) freeze 4.7 and Synapse repository.}
	\begin{tabular}{@{}llrr@{}} \toprule
		Platform & Input Data & \# of Genes  & \# of Samples \\\midrule
		P1. miRNA & syn2491366 & 14,345  &  4198\\
		P2. Methylation & syn2486658   & 1,383 & 4919 \\
		P3. Somatic CNV & syn1710678 & 876 &  3260\\
		P4. mRNA & syn1715755 & 14,178 &  3599\\
		P5. Somatic SNV & syn1729383 & 14,351 &  4933\\\bottomrule
	\end{tabular}
	\label{table:PanCan12}
\end{table}
\subsection{Data processing}
\subsubsection{Tensor construction with the PanCan12}\label{subsubsec:pancan12}
% data set description
Initially, Pan-Cancer-12 data freeze version 4.7 was downloaded from the Sage Bionetworks repository, Synapse \citep{Omberg2013}.
The PanCan12 contains multi-platform data with mapped clinical information of patients group into cohorts of twelve cancer type: bladder urothelial carcinoma (BLCA), breast adenocarcinoma (BRCA), colon and rectal carcinoma (COAD, READ), glioblastoma multiforme (GBM), head and neck squamous cell carcinoma (HNSC), kidney renal clear cell carcinoma (KIRC), acute myeloid leukaemia (LAML), lung adenocarcinoma (LUAD), lung squamous cell carcinoma (LUSC), ovarian serous carcinoma (OV), and uterine corpus endometrial carcinoma (UCEC).
Table~\ref{table:PanCan12} lists the Synapse IDs of the downloaded data for each platform used.
After download, probes of each platform are mapped to corresponding gene symbols.
Then, subjects and genes that have less than two evidences are removed from the dataset.
The resulting data for each platform are min-max normalized and is further normalized such that the Frobenius norm, i.e.,$\|A\|\equiv\sqrt{\sum_i\sum_j|a_{ij}|^2}$, becomes one.
A cell of resulting 3-mode tensor, $\T{X}$, contains a floating point value indexed on <$patient$, $gene$, $platform$> as shown in Figure~\ref{fig:overview}.
The size of the first mode spanning over subject or the patient index is 4,555; the size of the second mode spanning over the genes is 14,351; and the size of the third mode spans over five different platforms.
%\vspace{-.2cm}
%\vspace{-.5cm}

\subsubsection{Network constraint formation with pathway data}\label{subsubsec:pathway}
Links between genes in gene-gene network is used for constraining the factor matrices towards existing knowledge of gene associations.
The initial bio-network of human gene associations is retrieved from version 8 of PathwayCommons \citep{Cerami2011}.
The initial bio-network is then used to construct adjacency matrix of gene network for the list of gene considered in the tensor construction.
The adjacency matrix, $\mathbf{Y}$, contains 665,429 number of association information of 14,351 genes.

\subsection{Tensor basics}
We describe the basic notations and operations on a tensor and its decompositions.
Table \ref{table:symbols} shows the definitions of symbols used in this paper.
A tensor is a generalization of a multi-dimensional array denoted by a boldface Euler script, e.g. $\T{X}$.
An $N$-mode tensor is denoted as $\T{X}\in\mathbb{R}^{I_1\times I_2\times\cdots\times I_N}$, and ($i_1i_2\cdots i_N$)-th elements of $\T{X}$ is denoted as $x_{i_1i_2\cdots i_N}$.
A matrix is denoted by an uppercase bold letter, e.g. $\mathbf{A}$.
The $i$-th row vector of $\mathbf{A}$ is denoted by $\mathbf{a}_i$ in lowercase bold letter, and the ($ij$)-th entry of $\mathbf{A}$ is denoted by $a_{ij}$.
All tensor and matrix indices are positive integers greater than or equal to 1.
The mode-$n$ matrix product of a tensor $\T{X}\in\mathbb{R}^{I_1\times I_2\times\cdots\times I_N}$ with a matrix $\mathbf{A}\in\mathbb{R}^{I_n\times K}$ is denoted by $\T{X}\times_{n}\mathbf{A}$ and has the size of $I_1\times\cdots I_{n-1}\times K\times I_{n+1}\cdots\times I_N$. The element-wise definition is as follows:
%\vspace{-.2cm}
\begin{equation} \label{eqn:n mode matrix product elementwise}
(\T{X} \times_n \mathbf{A})_{i_1 \dots i_{n-1} j i_{n+1} \dots i_N} = \sum\limits_{i_n=1}^{I_n} x_{i_1 i_2 \dots i_N}a_{ji_n}.
%\vspace{-.1cm}
\end{equation}

See \cite{kolda2009tensor} for detailed explanations about tensor operations.
We focus on 3-mode tensor $\T{X}\in\mathbb{R}^{I_1\times I_2\times I_3}$ for the following sections since our dataset includes a 3-mode tensor (Section \ref{subsubsec:pancan12}).

After construction of tensors, they can be decomposed in several ways. We focus on higher order singular value decomposition (HOSVD).
HOSVD, also known as Tucker decomposition, is the generalization of singular value decomposition (SVD), which works on matrices.
HOSVD decomposes a tensor into a core tensor and orthogonal factor matrices corresponding to modes.
Specifically, given a 3-mode data tensor $\T{X}$, HOSVD decomposes $\T{X}$ as follows:
\begin{equation}\label{eqn:hosvd definition}
\T{X}\approx\tilde{\T{X}}=
\T{G}\times_{1} \mathbf{U}^{(1)} \times_2 \mathbf{U}^{(2)} \times_3 \mathbf{U}^{(3)},
%\vspace{-.1cm}
\end{equation}
where $\T{G} \in\mathbb{R}^{J_1\times J_2\times J_3}$ is a core tensor, and $\mathbf{U}^{(n)} \in\mathbb{R}^{I_n\times J_n}$ denotes the factor matrices for the $n$-th dimensions, respectively.
%More specifically, element-wise formulation of HOSVD is
%\begin{equation}\label{eqn:hosvd element}
%\begin{split}
%    x_{i_1i_2i_3}
%    \approx \tilde{x}_{i_1i_2i_3}&=\T{G}\times_{1} \mathbf{u}_{i_1}^{(1)} \times_2 \mathbf{u}_{i_2}^{(2)} \times_3 \mathbf{u}_{i_3}^{(3)}\\
%    &=\sum_{j_1=1}^{J_1}\sum_{j_2=1}^{J_2}\sum_{j_3=1}^{J_3}g_{j_1j_2j_3}u_{i_1j_1}^{(1)}u_{i_2j_2}^{(2)}u_{i_3j_3}^{(3)}.
%\end{split}
%\end{equation}
HOSVD finds the factors by minimizing the following objective function:
\begin{equation}\label{eqn:objective function}
\begin{split}
f&=\frac{1}{2}\|\T{X}-\tilde{\T{X}}\|^2+\frac{\lambda}{2} R(\T{G},\mathbf{U}^{(1)},\mathbf{U}^{(2)},\mathbf{U}^{(3)}),\\
&=\frac{1}{2}\sum_{(ijk)\in\Omega_{\T{X}}}(x_{ijk}-\T{G}\times_{1} \mathbf{u}^{(1)}_{i} \times_2 \mathbf{u}^{(2)}_{j} \times_3 \mathbf{u}^{(3)}_{k})^2\\
&+\frac{\lambda}{2} R(\T{G},\mathbf{U}^{(1)},\mathbf{U}^{(2)},\mathbf{U}^{(3)}),
\end{split}
%\vspace{-0.2cm}
\end{equation}
where $R(\T{G},\mathbf{U}^{(1)},\mathbf{U}^{(2)},\mathbf{U}^{(3)})$ is the $L_2$ regularization term.
Performance comparison of existing graph constrained HOSVD method is provided in the Supplementary.
%There are various existing optimization techniques for HOSVD including ALS and gradient-based approaches.
%Stochastic gradient descent (SGD) method (\cite{karatzoglou2010multiverse}) is appropriate for sparse tensor, highly scalable to large data and easily parallelizable to multiple cores, thus we adopt the SGD method for extracting the factors $\mathbf{U}$, $\mathbf{V}$, $\mathbf{W}$, and $\T{G}$.
\begin{table}[tbp]
	\caption{Table of symbols.}
	\begin{tabular}{@{}cl@{}}
		\toprule
		\textbf{Symbol} & \textbf{Definition}\\
		\midrule
		$\T{X}$ & a tensor (boldface Euler script)\\
		%            $\mathbf{X}_{(n)}$ & mode-$n$ matricization of a tensor\\
		$x_{ijk}$ & $(ijk)$-th entry of $\T{X}$\\
		$\mathbf{A}$ & a matrix (uppercase, bold letter)\\
		$\mathbf{a}_{i}$ & the $i$-th row vector of $\mathbf{A}$ (lowercase, bold letter)\\
		$a_{ij}$ & $(ij)$-th entry of $\mathbf{A}$\\
		$\times_n$ & $n$-mode matrix product\\
		$\|\bullet\|$ & Frobenius norm\\
		$\ast$ & Hadamard product \\
		$\circ$ & Outer product\\
		$\oslash$ & Element-wise division \\
		$\Omega_{\T{X}}$ & index set of $\T{X}$\\
		$\Omega_{\T{X}}^{n,i}$ & subset of $\Omega_{\T{X}}$ having $i$ as the $n$-th index\\
		$I_n$ & length of $n$-th dimension of input tensor $\T{X}$\\
		$J_n$ & length of $n$-th dimension of core tensor $\T{G}$\\
		\bottomrule
	\end{tabular}
	\label{table:symbols}
%	\vspace*{-0.5cm}
\end{table}

\begin{algorithm*} [!htb]
	\small
	\caption{SNeCT} \label{alg:netHOSVD}
	\begin{algorithmic}[1]
		\REQUIRE Input data: tensor $\T{X} \in \mathbb{R}^{I_1 \times I_2 \times \cdots \times I_N}$, network matrix $\mathbf{Y} \in \mathbb{R}^{I_c \times K}$, number of parallel cores $P$, and network-constrained mode $c$\\
		Hyperparameters: core size $(J_1, J_2, \cdots ,J_N)$, learning rate $\eta$, and regularization factors $\lambda$ and $\lambda_g$
		
		\ENSURE Core tensor $\T{G} \in \mathbb{R}^{J_1 \times J_2 \times \cdots \times J_N}$, and factor matrices $\mathbf{U}^{(1)}, \mathbf{U}^{(2)}, \cdots,\mathbf{U}^{(N)}$
		
		\STATE Initialize $\T{G}$, $\mathbf{U}^{(n)} \in \mathbb{R}^{I_n \times J_n}$ for $n= 1, 2, \cdots, N$ randomly
		
		\REPEAT
		
		\FORP{$\forall x_{i_1i_2 \cdots i_N}\in\T{X}$, $\forall y_{k_1k_2}\in{\mathbf{Y}}$ in random order}
		
		\IF{$x_{i_1i_2 \cdots i_N}\in\T{X}$ is picked}
		
		\STATE Cache intermediate data tensor: $\T{D} \leftarrow \T{G}\ast(\mathbf{u}^{(1)}_{i_1}\circ\mathbf{u}^{(2)}_{i_2}\circ\cdots\circ\mathbf{u}^{(N)}_{i_N}) $
		
		\STATE $\tilde{x}_{i_1i_2 \cdots i_N} \leftarrow $ sum of all elements of $\T{D}$
		
		\STATE Update corresponding factor rows: $\mathbf{u}^{(n)}_{i_n} \leftarrow \mathbf{u}^{(n)}_{i_n}-\eta\big((\tilde{x}_{i_1i_2 \cdots i_N}-x_{i_1i_2 \cdots i_N})\cdot
		Collapse(\T{D},n)+\frac{\lambda}{|\Omega_{\T{X}}^{n,i_n}|}\mathbf{u}^{(n)}_{i_n}
		\big)$, (for $n=1,2,\cdots,N$)
		
		\STATE Update core tensor: $\T{G} \leftarrow \T{G}-\eta P
		\big((\tilde{x}_{i_1i_2 \cdots i_N}-x_{i_1i_2 \cdots i_N})\cdot \T{D}\oslash\T{G}
		+\frac{\lambda}{|\Omega_{\T{X}}|}\T{G}
		\big)$
		, (executed by only one core)
		
		\ENDIF
		
		\IF{$y_{k_1k_2}\in\mathbf{Y}$ is picked}
		
		\STATE Update network-constrained factors: $\mathbf{u}^{(c)}_{k_1} \leftarrow \mathbf{u}^{(c)}_{k_1}-\eta\lambda_g y_{k_1k_2}
		(\mathbf{u}^{(c)}_{k_1}-\mathbf{u}^{(c)}_{k_2})
		$,~
		$\mathbf{u}^{(c)}_{k_2} \leftarrow \mathbf{u}^{(c)}_{k_2}-\eta\lambda_g y_{k_1k_2}
		(\mathbf{u}^{(c)}_{k_2}-\mathbf{u}^{(c)}_{k_1})
		$
		
		\ENDIF
		
		\ENDFOR
		
		\UNTIL{convergence conditions are satisfied}
		
		\STATE $\mathbf{Q}^{(n)}$, $\mathbf{R}^{(n)} \leftarrow$ QR decomposition of $\mathbf{U}^{(n)}$, ~
		$\mathbf{U}^{(n)} \leftarrow \mathbf{Q}^{(n)}$
		, $\T{G}\leftarrow\T{G}\times_{n}\mathbf{R}^{(n)}$, (for $n=1, 2, \cdots, N$)
		
		\RETURN $\T{G}$, $\mathbf{U}^{(1)}, \mathbf{U}^{(2)}, \cdots,\mathbf{U}^{(N)}$
		
	\end{algorithmic}
\end{algorithm*}

\subsection{SNeCT decomposition}

\subsubsection{HOSVD optimization with network constraint}
Consider that a graph $\mathit{G}$ represents a network between entities of a dimension. For our data tensor with dimensions of (patient, gene, platform), $\mathit{G}$ is the network of genes as explained in Section \ref{subsubsec:pathway}.
$\mathit{G}$ informs the similarities between genes, e.g., gene $i$ and gene $j$ are similar if $y_{ij}=1$.
To include the similarity constraint to HOSVD, the network graph $\mathit{G}$ acts as a regularization as studied in previous works of \cite{narita2012tensor,li2009relation}.
Specifically, we add the network regularization term $\lambda_g f_g$ to the objective function of
Equation \eqref{eqn:objective function}
where matrix $\mathbf{Y}$ is the adjacency matrix of $\mathit{G}$ constraining the second dimension, and $\lambda_g$ is a constant.
%\begin{equation}\label{eqn:graph regularization}
%f_g:=tr(\mathbf{U}^{(2)\mathsf{T}}\mathbf{L}\mathbf{U}^{(2)})
%\end{equation}
\begin{equation}
\label{eqn:graph regularization}
\begin{split}
f_g&= \frac{1}{2} \sum_{l=1}^{J_2}\Big[\sum_{(k_1k_2)\in\Omega_{\mathbf{Y}}}y_{k_1k_2}(u^{(2)}_{k_1l}-u^{(2)}_{k_2l})^2\Big]\\
&= \frac{1}{2} \sum_{(k_1k_2)\in\Omega_{\mathbf{Y}}}y_{k_1k_2}\|\mathbf{u}^{(2)}_{k_1}-\mathbf{u}^{(2)}_{k_2}\|^2\\
\end{split}
\end{equation}
where %$tr(\cdot)$ denotes the trace of a matrix, and
$\Omega_{\mathbf{Y}}$ is the index set of $\mathbf{Y}$.
%\blue{When $\mathbf{Y}$ is an undirected graph, we take only the upper triangular part of $\mathbf{Y}$ as input to prevent duplicate updates.}
Minimizing $f_g$ guides the algorithm such that the factors of associated genes become similar, i.e., $\mathbf{u}^{(2)}_{j_1}$ and $\mathbf{u}^{(2)}_{j_2}$ have similar values when there is an edge between gene $j_1$ and gene $j_2$ in the graph $\mathit{G}$.

\subsubsection{Parallelizable update rules}
We present a multi-core algorithm to minimize the objective function $f_{opt}=f+\lambda_gf_g$ and factorize the given tensor $\T{X}$ into HOSVD form.
SNeCT adopts parallel stochastic gradient descent (SGD) optimization technique, and thus is highly memory-efficient and scalable to large datasets and multiple cores.
We rewrite $f$ so that it forms an SGD-amenable form.
\begin{equation*}
f= \frac{1}{2}\sum\limits_{(i_1i_2i_3)\in\Omega_{\T{X}}}\Big[\big(x_{i_1i_2i_3}-\tilde{x}_{i_1i_2i_3}\big)^2
+\frac{\lambda}{|\Omega_{\T{X}}|}\|\T{G}\|^2
+\lambda\sum_{n=1}^{3}\frac{\|\mathbf{u}_{i_n}^{(n)}\|^2}{|\Omega_{\T{X}}^{n,i_n}|}\Big]
\end{equation*}
Note that $f_g$ already has the SGD-amenable form and $\Omega_{\T{X}}$ excludes the cells with missing data. Now, gradients of $f_{opt}$ with respect to factors for a given data point $x_{\alpha=(i_1i_2i_3)}$ or $y_{\beta=(k_1k_2)}$ are calculated as follows:

\begin{equation}\label{eqn:gradient}
\begin{split}
\left.\frac{\partial f_{opt}}{\partial \mathbf{u}^{(1)}_{i_1}}\right\vert_{\alpha}&=
-\big(x_{\alpha}-\tilde{x}_{\alpha}\big)
\big{\lbrack}\T{G}\times_{2}\mathbf{u}^{(2)}_{(i_2)}\times_{3}\mathbf{u}^{(3)}_{(i_3)}
\big{\rbrack}
+\frac{\lambda}{|\Omega_{\T{X}}^{1,i_1}|}\mathbf{u}^{(1)}_{i_1}\\
\left.\frac{\partial f_{opt}}{\partial \T{G}}\right\vert_{\alpha}&=
-\big(x_{\alpha}-\tilde{x}_{\alpha}\big)
\times_1\mathbf{u}^{(1)\mathsf{T}}_{i_1}
\times_2\mathbf{u}^{(2)\mathsf{T}}_{i_2}
\times_3\mathbf{u}^{(3)\mathsf{T}}_{i_3}
+\frac{\lambda}{|\Omega_{\T{X}}|}\T{G}\\
\left.\frac{\partial f_{opt}}{\partial \mathbf{u}^{(2)}_{k_1}}\right\vert_{\beta}&=
\lambda_gy_{\beta}(\mathbf{u}^{(2)}_{k_1}-\mathbf{u}^{(2)}_{k_2})
\end{split}
\end{equation}
$\left.\frac{\partial f_{opt}}{\partial \mathbf{u}^{(2)}_{i_2}}\right\vert_{\alpha}$,
$\left.\frac{\partial f_{opt}}{\partial \mathbf{u}^{(3)}_{i_3}}\right\vert_{\alpha}$, and
$\left.\frac{\partial f_{opt}}{\partial \mathbf{u}^{(2)}_{k_2}}\right\vert_{\beta}$
are calculated symmetrically as the above equations. 
The above equations are naturally generalized to mode-$N$ tensors.

\subsubsection{SNeCT learning model}
SNeCT optimizes the objective function $f_{opt}$ by parallel SGD update. Algorithm \ref{alg:netHOSVD} shows detailed procedures of decomposition of a general $N$-mode tensor $\T{X}$ and network constraint $\mathbf{Y}$ which represents the similarity of $c$-th mode entities.

In the beginning, SNeCT initializes $\mathbf{U}^{(1)}$, $\mathbf{U}^{(2)}$, $\cdots$, $\mathbf{U}^{(N)}$, and $\T{G}$ randomly (line 1 of Algorithm \ref{alg:netHOSVD}). The outer loop (lines 2-14) repeats until the factors converge.
In the inner loop (lines 3-13), SNeCT conducts parallel updates of factor rows corresponding to each data point $x_{i_1i_2i_3}$ or $y_{j_1j_2}$ in random order.
When calculating gradients with respect to factor rows and core tensor,
it takes excessive time to calculate $\tilde{x}_{alpha}=\T{G}\times_1\mathbf{u}^{(1)}_{i_1}\times_2\cdots\times_n\mathbf{u}^{(N)}_{i_N}$ and tensor-matrix products for $\frac{\partial f_{opt}}{\partial \mathbf{u}^{(n)}_{i_n}}$ every time when they are needed.
SNeCT reduces the time cost efficiently by caching intermediate data tensor $\T{D}$ (line 5). See section 3.5 of \cite{choi2017fast} for detailed approach of utilizing intermediate data to reduce time cost.
$Collapse(\T{D},n)$ operator (line 7) outputs a vector with length of $i_n$ which contains the sum of $k$-th slice of $D$ over $n$-th mode as its $k$-th element.
In line 8, element-wise division operator $\oslash$ is used to efficiently calculate core tensor gradient.

There are possible conflicts between the parallel updates since a factor row or core tensor might be accessed by multiple update attempts.
However, we apply lock-free parallel update scheme \citep{recht2011hogwild, choi2017fast} and remove frequent conflicts by updating core tensor using only one core (line 6); thus SNeCT guarantees near-linear convergence to a local optimum.

\section{Results}
    \label{sec:exp}
    \subsection{Stratification}
\subsubsection{Cluster assignment}\label{subsubsec:coc}
We perform cluster analysis with $k$-means clustering algorithm on patient profiles using Euclidean distance.
(Euclidean, cosine and Mahalanobis distance measures have been tested with no significant differences in the results.)
To generate patient profiles, SNeCT decompose the data tensor as shown in Fig.~\ref{fig:overview} with core size [$J_1, J_2, J_3$] = [78, 48, 5], where the best core size was searched with a small validation tensor and graph constraint is set to $\lambda=1.0$.
After decomposition, the rows of patient factor matrix $U^{(1)}$ is used as patient profiles.
To find the cluster size we compute the gap statistics introduced by \cite{Tibshirani2001}.
The gap statistics of the cluster stabilizes after cluster size of 10 as shown in Supplementary Fig.~3%\ref{fig:gapStat}.
For convenience of comparison, we stratified patient profiles into 13 clusters as suggested in \cite{Hoadley2014}.
Table \ref{tab:COCA} shows the number of assigned patients of twelve cancer types for each cluster.
\begin{table}[!htbp]
	\small
	\caption{12 pathological disease types assigned to clusters of profiles factorized from tensor factorization with graph constraint of $\lambda = 1.0$.}
	\begin{tabular}{@{}l@{ }r@{ }r@{ }r@{ }r@{ }r@{ }r@{ }r@{ }r@{ }r@{ }r@{ }r@{ }r@{ }r@{ }r@{}}
		\toprule
		& \textbf{C1} & \textbf{C2} & \textbf{C3} & \textbf{C4} & \textbf{C5} & \textbf{C6} & \textbf{C7} & \textbf{C8} & \textbf{C9} & \textbf{C10} & \textbf{C11} & \textbf{C12} & \textbf{C13} & \textbf{Total} \\
		\midrule
		BLCA & 16 & 32 & 2 & 19 & 0 & 22 & 3 & 0 & 0 & 0 & 32 & 0 & 0 & 126\\
		BRCA & 17 & 3 & 600 & 172 & 1 & 70 & 0 & 0 & 0 & 0 & 26 & 0 & 0 & 889\\
		COAD & 4 & 0 & 2 & 2 & 0 & 91 & 317 & 0 & 0 & 0 & 1 & 2 & 0 & 419\\
		GBM & 4 & 1 & 1 & 2 & 3 & 7 & 0 & 0 & 248 & 0 & 1 & 0 & 0 & 267\\
		HNSC & 0 & 242 & 1 & 6 & 0 & 1 & 0 & 0 & 0 & 0 & 60 & 0 & 0 & 310\\
		KIRC & 14 & 1 & 1 & 0 & 471 & 4 & 0 & 0 & 1 & 0 & 6 & 0 & 0 & 498\\
		LAML & 0 & 0 & 0 & 0 & 0 & 9 & 0 & 0 & 0 & 188 & 0 & 0 & 0 & 197\\
		LUAD & 302 & 2 & 2 & 7 & 1 & 12 & 0 & 0 & 0 & 0 & 29 & 0 & 0 & 357\\
		LUSC & 26 & 32 & 0 & 29 & 0 & 7 & 0 & 0 & 0 & 0 & 246 & 0 & 0 & 340\\
		OV & 0 & 0 & 1 & 3 & 0 & 1 & 1 & 348 & 0 & 0 & 0 & 0 & 131 & 485\\
		READ & 1 & 1 & 0 & 5 & 0 & 9 & 145 & 0 & 0 & 0 & 1 & 1 & 0 & 163\\
		UCEC & 3 & 1 & 3 & 117 & 1 & 348 & 1 & 0 & 0 & 0 & 10 & 13 & 2 & 499\\
		Total & 387 & 315 & 613 & 362 & 477 & 581 & 467 & 348 & 249 & 188 & 412 & 17 & 134 & 4550 \\
		\bottomrule
	\end{tabular}
	
	\label{tab:COCA}
%	\vspace*{-0.5cm}
\end{table}

\subsubsection{Survival analysis}
\begin{figure} [!htbp]
	%\subfloat[$\lambda_g=0$]{\includegraphics[width=0.16 \textwidth]{FIG/KM_plot_without_Net.eps}}
	%\subfloat[$\lambda_g=0.1$]{\includegraphics[width=0.16 \textwidth]{FIG/KM_plot_with_Net01.eps}}
	%\subfloat[$\lambda_g=1$]{\includegraphics[width=0.16 \textwidth]{FIG/KM_plot_with_Net.eps}}
	\includegraphics[width=0.45\textwidth]{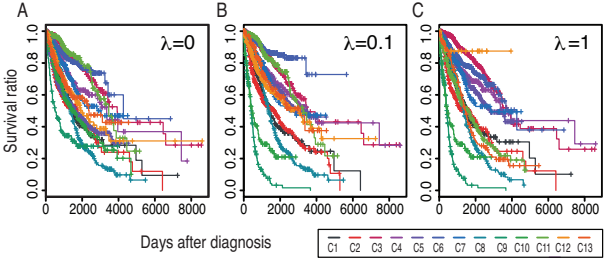}
%	\vspace*{-0.2cm}
	\caption{Predicted survival curves for clustered patients. x-axis is survival time (day) and y-axis is survival rate.}
	\label{fig:survAnalysis}
%	\vspace*{-0.5cm}
\end{figure}

We performed survival analysis for the thirteen clusters acquired from Section \ref{subsubsec:coc}, using the Cox proportional hazards regression model in the R survival package (Fig~\ref{fig:survAnalysis}).
We use right-censored survival data for patients: days to death for dead patients, and days to last contact for alive patients as right-censored data.
To see how the network constraint affects decomposition result, we impose three different levels of network regularization: $\lambda_g$ value of 0 (not constrained), 0.1, and 1.
The log-rank statistics value for $\lambda_g=0$ is 409, for $\lambda_g=0.1$  is 1151, and for $\lambda_g=1$ is 1185.
The Figure~\ref{fig:survAnalysis} and the log-rank statistics values show that having graph constraint is better, however, little difference is observed for the two weight values.
%\subsubsection{Platform factor analysis}
%We analyze the degree of influence for each platform (Table \ref{table:PanCan12}).
%Norm of a slice of $\T{G}\times_{3}\mathbf{U}^{(3)}$ corresponding to each platform is shown in Figure \ref{fig:platformNorm}.
%Note that P1(miRNA) is the most dominant, and P4(mRNA) is the next dominant platform for our dataset.
\subsubsection{Unique cohort clusters}
Cluster result of the patient factor matrix, $\mathbf{U}^{(1)}$, correlates with tissue of origin which is similar to observation made by \cite{Hoadley2014}.
Six clusters, C1-LUAD-enriched, C3-BRCA/Luminal, C5-KIRC, C8-OV-1, C9-GBM, C10-LAML, and C13-OV-2 each dominantly contains cancer samples from a single tissue of origin.
Patients in C1-LUAD-enriched cluster includes 302 out of 357 LUAD patients with relatively good prognosis, that is, neoplasm cancer status is tumor free for 186 out of all 217 tumor free cases with precision of 0.73 (recall of 0.85).
Patients in C3-BRCA/Luminal cluster groups 600 BRCA cases with 13 other cancer types.
The BRCA patients in C3 have positive estrogen and progesterone receptor status with precision of 0.95 and 0.85, respectively, and HER2 status is mixed tending to have more negative status with precision of 0.73.
Furthermore, C3 contains 8 out of 9 metastatic cases and contains 34 out of 43 cases with known other malignancy histological type.
It tells us that C3 groups patients with BRCA Luminal A and Luminal B molecular subtypes of breast cancer.
Four other clusters that form somewhat mutually exclusive collectively exhaustive groups are cluster C5-KIRC that contains 471 patients classified as KIRC and 6  patients classified to other cancer types, cluster C9-GBM that contains 248 cases of GBM patients with only one of KIRC patient, cluster C10-LAML that contains 188 cases of LAML with no other cancer patients included, and cluster C12-UCEC-small, a small cluster, that contains 13 UCEC with 4 other cases with very high survival ratio as shown in  Fig.~\ref{fig:OV_BRCA_UCEC}(c).

Interestingly, two clusters are formed for OV cases: C13-OV-2 that contains 131 cases of OV patient with only three other cancer types and cluster C8-OV-1 that contains 348 cases of OV patients with no other cancer type.
There is no clear distinction between the clinical features of the two clusters.
However, a clear distinction can be found comparing factor values of genomic contents for C13, C8, and other clusters, as shown in Supplementary Fig.~4(a).%\ref{fig:OV}(a).
Separation can also be found in the survival analysis of C8 and C13 (Fig.~\ref{fig:OV_BRCA_UCEC}(a)).

Another interesting cluster is C7-COAD/READ that combines two colorectal tumor samples from cohorts COAD and READ with C7 READ containing almost all READ cases (145 out of 163).
C7 COAD group contains 22 (precision 1 and recall of 0.88) normal braf gene analysis results while C6 COAD contains all three cases of abnormal braf gene analysis result.
C7 COAD tends to have better survival ratio compared to C6 COAD cases as shown in Supplementary Fig.~5.%\ref{fig:others}.

\subsubsection{Squamous-like cluster}
\begin{figure} [!htbp]
	%    \subfloat[OV clusters]{\includegraphics[width=0.16\textwidth]{FIG/KM_plot_with_Net_cancer_OV.eps}}
	%    \subfloat[BRCA clusters]{\includegraphics[width=0.16\textwidth]{FIG/KM_plot_with_Net_cancer_BRCA.eps}}
	%    \subfloat[UCEC clusters]{\includegraphics[width=0.16\textwidth]{FIG/KM_plot_with_Net_cancer_UCEC.eps}}
	\includegraphics[width=0.45\textwidth]{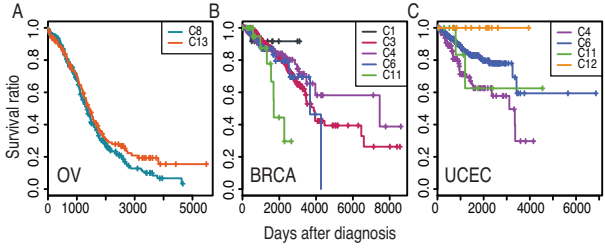}
%	\vspace*{-0.2cm}
	\caption{Survival analysis of clusters for cohorts OV (A), BRCA (B), and UCEC (C).}
	\label{fig:OV_BRCA_UCEC}
%	\vspace*{-0.5cm}
\end{figure}

Patients in the C2-HNSC-enriched-squamous-like cluster contains squamous-like BLCA (32), HNSC (242), and LUSC (32) of mostly male (228/315) patients.
242 out of 310 HNSC patients are group to C2 where 60 other HNSC patients are grouped in another squamous-like cluster C11 that contains 246 out of 340 LUSC patients.
Also, within C2 cluster, BLCA group contains 26 patients diagnosed as non-papillary with a precision of 0.79 and mixed neoplasm cancer status (14 tumor free and 13 with tumor) and contains 9 cases of having other malignancy histological type.
LUSC and HNSC patients in group C2 do not show clear characteristics in terms of clinical features other then being skewed towards male gender.
%However, LUSC group are does not show clear characteristics having mixed neoplasm cancer status of 19 tumor free and 5 with tumor.
%"BLCA: The SCC-like subtype was characterized by high expression of basal keratins normally not expressed in the urothelium, KRT4, KRT6A, KRT6B, KRT6C, KRT14, and KRT16, as well as by bad prognosis. As these keratins have been associated with squamous differentiation of urothelial cell carcinoma \citep{Sjodahl2012}"
Cluster C11-LUSC-enriched-squamous-like contains a mixed group of 32 BLCA, 26 BRCA, 60 HNSC,  29 LUAD, and 246 LUSC cancer patients.
The LUSC cases within C11 contains 128 patients and tends to be skewed towards patients with higher level of smoking history (level 4 - 130/174, level 3 - 43/67, level 2 61/79 and level 1 - 6/13) but with slightly better prognosis (tumor free 142 out of 198 and with tumor 38 out of 56);
%BLCA patients in C11 tend to have mixed papillary diagnosis subtypes (14 papillary 16 non-papillary).
%BRCA:  24 cases are labeled the icd-o-3 histology of 8500/3;
HNSC patients in C11 have mixed neoplasm cancer statuses (tumor free 36 and with tumor 18) and are mostly of intermediate histological grade (41 of grade g2 and 15 of grade g3).
BRCA patients in the group tend to have negative progesterone receptor status (20 negatives; 4 positives), negative estrogen receptor status (16 negatives; 8 positives), negative  HER2 status (14 negatives; 3 positives; 3 equivocals), contains many (24) infiltrating ductal carcinoma cases, and tend to have the worst prognostics (Fig.~\ref{fig:OV_BRCA_UCEC}(a)).
%LUAD: 24 has the icd-10 code c34.1 (precision 0.8) and 5 has code c34.3. 23 cases with the icd-o-3 histology of 8140/3 (precision 0.77)
\subsubsection{BRCA/UCEC clusters}
Two BRCA/UCEC-enriched clusters are formed as C4 and C6.
C4-BRCA/UCEC-enriched cluster contains 172 BRCA and 117 UCEC cases mixed with 29 LUSC and 19 BLCA cases.
C6-BRCA/UCEC-enriched cluster contains 70 BRCA and  348 UCEC cases mixed with 90 COAD and 22 BLCA cases.
% BRCA
Patients in C4 BRCA has mostly negative estrogen receptor status (127 negative cases with precision 0.77; 38 positives), mostly negative progesterone receptor status (131 negatives; 32 positives; 1 intermediate), and mostly negative her2 status(101 negatives; 23 positives; 23 equivocals).
The group also contains all 5 of medullary carcinoma cases.
70 BRCA patients in C6, on the other hand, have mixed positive and negative status for both progesterone receptor estrogen receptor and mostly negative HER2 status (38 negatives; 8 positives; 9 others).
Also, survival analysis shows higher ratio of patients surviving longer for C4-BRCA compared to C6-BRCA.
%UCEC
The clinical characteristics and the survival analysis show that C4 UCEC groups patients with poor prognostic.
That is, C4 UCEC includes higher ratio of serous adenocarcinoma (69/115) cases than the common endometrioid carcinoma (45/378); the tumor grades of the group tend to be high with 7 high-grades (out of 11 possible) and 100 grade-3; 12 out of 30 cases are with known as `other malignancy histological type' within the group; and the group shows the fastest drop of survival rates in the Kaplan-Meier plot (Fig.~\ref{fig:OV_BRCA_UCEC}(b).)
On the other hand, UCEC patients grouped in C6 show better prognostic.
That is, the C6 UCEC includes large portion of endometrioid carcinoma (312/375, precision of 0.89), large portion of tumor free neoplasm cancer status (292/393, precision of 0.88), large portion of low to intermediate histological grades (g1 - 87/91, g2 - 93/106), and approximately half of grade 3 and little of grade 4 (3 out of 11).
Also, their survival ratio drop rate is less dramatic compared to C4 UCEC cases.
% C4:
%LUSC: no strong tendency was found. various smoking history and histology and other labels. (check if mixed tumor)
%BLCA: 18 of 19 has the icd o 3 histology  8120/3 and one out of one 8070/3; all 19 has histological type of muscle-invasive urothelial carcinoma (pt2 or above); contains more non-papillary diagnosed cases (16 non-papillary vs 3 papillary); all 19 has high neoplasm histologic grade; 7 of which is know to have other malignancy histological type associated;  - bad prognosis (?)
% C6 contains
%22 BLCA patients contain 15 of the icd 10 code of c67.9; the icd-o 3 histology 19 of $8120/3$ and 3 of $8130/3$; with no
%We use the ground truth data of patients on which cancer they have among bladder urothelial carcinoma (BLCA), breast adenocarcinoma (BRCA), colon and rectal carcinoma (COAD, READ), glioblastoma multiforme (GBM), head and neck squamous cell carcinoma (HNSC), kidney renal clear cell carcinoma (KIRC), acute myeloid leukaemia (LAML; conventionally called AML), lung adenocarcinoma (LUAD), lung squamous cell carcinoma (LUSC), ovarian serous carcinoma (OV), and uterine corpus endometrial carcinoma (UCEC).
%\vspace*{-0.2cm}

\subsection{Prediction}
\subsubsection{Top-$k$ search}
When a new query patient $q$ arrives with data $\T{X}_{q}$ which is a tensor representing the patient profile of $q$, we aim to find the SNeCT factor for the patient using the pre-calculated factor matrices and core tensor.
Thus we compare the factors of other patients which are encoded in the patient factor matrix $\mathbf{U}^{(1)}$ with the calculated patient factor.
We solve the following equation with the SNeCT algorithm while fixing parameters other than $\mathbf{u}$.
\begin{equation}\label{eqn:new}
\mathbf{u}_q=\arg\min_{\mathbf{u}}\|\T{X}_{q}-\T{G}\times_1 \mathbf{u} \times_2 \mathbf{U}^{(2)} \times_3 \mathbf{U}^{(3)}\|.
\end{equation}
After generation of new profile, $\mathbf{u}_q$, it is used to seek top-$k$ similar patients by calculating the distance between the query factor and patient factors.
It takes $550$ms on average to search for a query against the training set.

\subsubsection{Clinical similarity of top-$k$ search}
% SNeCT finds the query profile vector $\mathbf{u}_q$ when query data for a new patient arrives.
To test the clinical prediction accuracy of SNeCT, we generated the factor matrices from 90\% of the data and used 10\% of the data as test set or new queries and determined the clinical similarity of the query patient to the top-$k$ similar patients searched against factor matrix $\mathbf{U}^{(1)}$.
Then we evaluated the average precision over test cases for each selected clinical features on top-1, top-5, top-10, and top-$R$ search results, where the $R$ value computes the number of samples with the same clinical values as the query in the database and varies from query to query.
Overall, ``age at initial pathologic diagnosis'' and ``vital status'' coincide well with all top-$k$ retrievals with average precision over all the test data ranging from 0.76 to 0.81 and from 0.66 to 0.68, respectively.
No significant features were found for LUAD, LUSC, and BLCA other than the two clinical features.
Other clinical features that are cohort-specific or have high average precision values are listed in Table~\ref{tab:topk}.
Looking at the precision values, we can see that the search successfully retrieved BRCA patients with similar estrogen and progesterone receptor status in most cases while less so in terms of her2/neu IHC receptor status.
Also, most search results matched that of the query for the braf gene analysis results in the COAD and READ test cases.
%\vspace*{-10pt}

\begin{table}[!tbp]
	\caption{Top-$k$ search precision.}
	\small
		\begin{tabular}{@{}ll@{}llll@{}}
			\toprule
			Cohort & Clinical Features            & Top 1 & Top 5 & Top 10 & Top R \\
			\midrule
			BRCA & estrogen receptor status       & 0.72  & 0.85  & 0.86   & 0.81  \\
			& progesterone receptor status   & 0.86  & 0.71  & 0.71   & 0.68  \\
			& her2/neu IHC receptor status   & 0.53  & 0.51  & 0.49   & 0.55  \\
			& neoplasm cancer status         & 0.84  & 0.80  & 0.81   & 0.77  \\
			COAD & braf gene analysis result      & 1.00  & 0.80  & 0.70   & 0.92  \\
			& colon polyps present           & 0.47  & 0.56  & 0.52   & 0.59  \\
			& 1st relatives with cancer      & 0.84  & 0.78  & 0.75   & 0.84  \\
			& venous invasion                & 0.61  & 0.60  & 0.63   & 0.72  \\
			GBM  & histological type              & 0.96  & 0.94  & 0.94   & 0.78  \\
			& icd-o-3 histology              & 1.00  & 1.00  & 1.00   & 0.77  \\
			& neoplasm cancer status         & 0.85  & 0.82  & 0.83   & 0.77  \\
			HNSC & hpv status by p16 testing      & 0.78  & 0.78  & 0.77   & 0.73  \\
			KIRC & histological type              & 1.00  & 0.99  & 0.99   & 0.73  \\
			& icd-o-3 histology              & 1.00  & 0.99  & 0.99   & 0.73  \\
			& number packs/year smoked       & 0.50  & 0.30  & 0.20   & 1.00  \\
			LAML & calgb cytogenetics risk cat.   & 0.85  & 0.84  & 0.81   & 0.65  \\
			OV   & neoplasm histologic grade      & 0.79  & 0.75  & 0.76   & 0.77  \\
			READ & braf gene analysis result      & 1.00  & 1.00  & 1.00   & 1.00  \\
			& 1st relatives with cancer      & 0.93  & 0.77  & 0.78   & 0.86  \\
			UCEC & menopause status               & 0.71  & 0.76  & 0.76   & 0.77  \\
			& neoplasm cancer status         & 0.83  & 0.75  & 0.77   & 0.77  \\
			\bottomrule
		\end{tabular}
	\label{tab:topk}
%	\vspace*{-0.5cm}
\end{table}

%\vspace*{-8pt}

\begin{figure*} [!htbp]
	\includegraphics[width=1\textwidth]{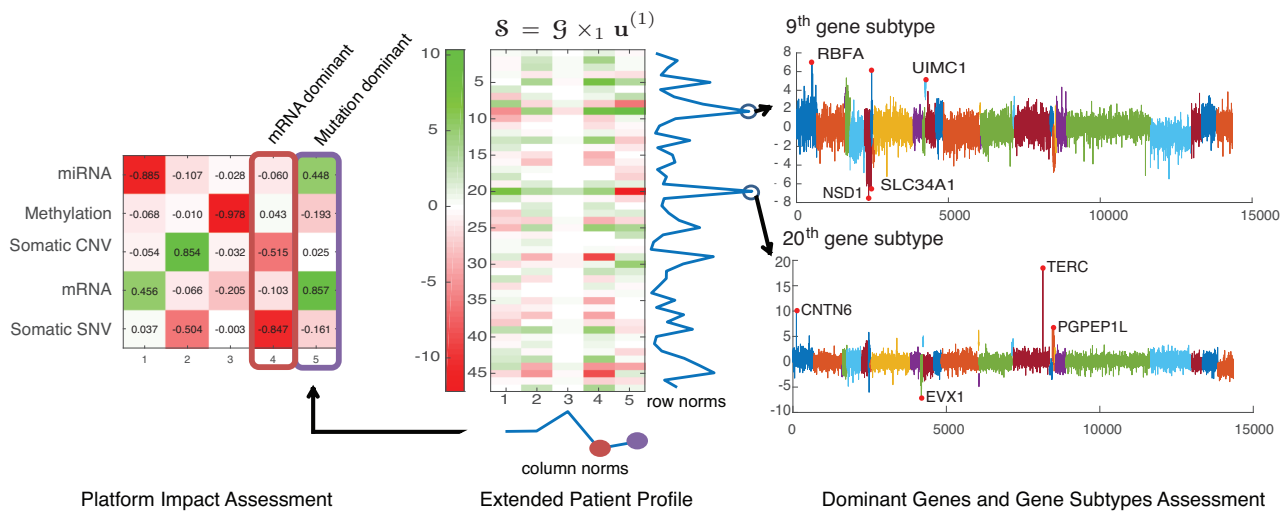}
%	\vspace*{-0.2cm}
	\caption{Personalized analysis example of patient ID TCGA-BS-A0UV in cohort UCEC assigned to cluster C12. }
	\label{fig:personalized}
%	\vspace*{-0.5cm}
\end{figure*}

\subsection{Personalized subtype analysis scenario}
The most valuable aspects of tensor factorization results are on the possibility of personalized interpretation of given patients.
To illustrate how factor matrices can be used for personalized interpretation, we provide a brief example on a given patient $i$.
For the patient $i$, SNeCT generates patient profile $\mathbf{u}^{(1)}_i$. If the patient is a new patient we can use the Eq.~\ref{eqn:new} to generate the profile.
We then calculate the personalized subtype matrix as follows: $\T{S}=\T{G}\times_1\mathbf{u}^{(1)}_i$ ($\in \mathbb{R}^{J_2 \times J_3}$).
$\T{S}$ provides a personalized weight information for subtypes for the gene and the platform modes.
For the sample TCGA-BS-A0UV, the center of Fig.~\ref{fig:personalized} shows the heatmap of $\T{S}$.
Each row of $\T{S}$ represents a subtype for gene mode, thus norm of each row represents the influence of each subtype to the patient.
%Figure \ref{fig:geneNorms} shows the calculated subtype weights for 13 representative patients for clusters from COCA (Section \ref{subsubsec:coc}).
Each column of $\T{S}\times_{3}\mathbf{U}^{(3)}$ represents the platform mode.
The norm of each column shows the influence of each platform to the patient and right side of Fig.~\ref{fig:personalized} shows the associated factor values.
With the analysis, we can determine which gene subtypes dominantly characterize the patient and which platform data were important in finding the dominant characteristics such that we can trace back to the significant genes and the type of abnormality.

%\vspace*{-10pt} 

%\section{Related Works}
%    \label{sec:related}
%    \input{050related}

\section{Discussion and Conclusions}
    \label{sec:concl}
    In this paper, we have proposed a large-scale network constrained Tucker decomposition method (SNeCT) that is based on parallelizable stochastic gradient descent.
With SNeCT, it is possible to systematically analyze high dimensional multi-platform genomic data constrained on prior knowledge of feature associations in a form of a network.
It is a general purpose approach that can be applied in various combinations of multi-platform data.
This is important as the availability and variety of multi-platform genomic data increases and the need for fast and intuitive methods becomes higher.
However, existing methods either run in a small-scale analysis, combine multiple analysis methods thus requiring a large number of hyper-parameter tuning and expert knowledge.
The practicality of SNeCT was shown on the PANCAN12 dataset where the stratification result shows a high correlation to subsets of clinical features and to the tissue of origin, which is consistent with the observation made by \cite{Hoadley2014}.
Also, SNeCT can be applied to search for top-$K$ similar patients given a new patient, which has various utilities such as using the clinical features of top-$k$ patients in diagnostics and prognostic predictions.
Furthermore, we showed how the combination of factor matrices can be used for personalized genomic interpretation of a patient.
There are considerations to make when using SNeCT, such as choosing normalization and gene mapping methods for construction of the input tensor, sizes of latent factors, appropriate network, number of clusters in stratification studies, and value of $K$ in the top-$K$ search in the clinical predictions.
However, these are common problems in analysis and several solutions can be found in existing literature.
We conclude that SNeCT provides a powerful tool for integrative analysis of multi-platform to the bioinformatics community.

\section*{Funding}
This research was supported by Basic Science Research Program through the NRF of Korea (NRF-2015R1C1A2A01055739) and by the KEIT Korea under the ``Global Advanced Technology Center'' (10053204).

\bibliographystyle{ACM-Reference-Format}

%\bibliography{BIB/other}

%\newpage
%\appendix
%\section{Appendix}
%\label{sec:appendix}
%\input{099appendix}

\end{document}